\documentclass[letterpaper, 10 pt, conference]{ieeeconf}  
\usepackage{algorithm2e}
\RestyleAlgo{ruled}

\IEEEoverridecommandlockouts                              

\let\labelindent\relax
\usepackage{enumitem}  
\usepackage{booktabs}  
\usepackage{multirow}  
\usepackage{graphicx}  
\usepackage{color}     
\usepackage{amssymb}
\usepackage{amsmath}

\newcommand{\bfa}{\mathbf{a}}
\newcommand{\bfo}{\mathbf{o}}

\newcommand{\ie}{i.e., }
\newcommand{\eg}{e.g., }

\title{\LARGE \bf
Implicit Kinematic Policies: Unifying Joint and Cartesian Action Spaces in End-to-End Robot Learning
}

\author{
Aditya Ganapathi$^{1,2}$,
Pete Florence$^1$,
Jake Varley$^1$,
Kaylee Burns$^{1,3}$,
Ken Goldberg$^2$,
Andy Zeng$^1$
}

\begin{document}

\maketitle
\thispagestyle{empty}
\pagestyle{empty}

\begin{abstract}

Action representation is an important yet often overlooked aspect in end-to-end robot learning with deep networks. Choosing one action space over another (e.g. target joint positions, or Cartesian end-effector poses) can result in surprisingly stark performance differences between various downstream tasks -- and as a result, considerable research has been devoted to finding the right action space for a given application. However, in this work, we instead investigate how our models can discover and learn for themselves which action space to use. Leveraging recent work on implicit behavioral cloning, which takes both observations and actions as input, we demonstrate that it is possible to present the same action in multiple different spaces to the same policy -- allowing it to learn inductive patterns from each space. Specifically, we study the benefits of combining Cartesian and joint action spaces in the context of learning manipulation skills. To this end, we present Implicit Kinematic Policies (IKP), which incorporates the kinematic chain as a differentiable module within the deep network. Quantitative experiments across several simulated continuous control tasks---from scooping piles of small objects, to lifting boxes with elbows, to precise block insertion with miscalibrated robots---suggest IKP not only learns complex prehensile and non-prehensile manipulation from pixels better than baseline alternatives, but also can learn to compensate for small joint encoder offset errors. Finally, we also run qualitative experiments on a real UR5e to demonstrate the feasibility of our algorithm on a physical robotic system with real data. See https://tinyurl.com/4wz3nf86 for code and supplementary material.







{\let\thefootnote\relax\footnote{{$^1$Robotics at Google, $^2$AUTOLab at UC Berkeley, $^3$Stanford University}}} 

\end{abstract}

\vspace{-1em}
\section{Introduction}

Deep visuomotor policies that map from pixels to actions end-to-end can represent complex manipulation skills \cite{levine_finn_2016}, but have shown to be sensitive to the choice of action space \cite{martin2019iros} -- e.g., Cartesian end effector actions (i.e. task space \cite{berenson2011task}) perform favorably when learning policies for tabletop manipulation, while joint actions have shown to fare better for whole-body motion control \cite{peng2017learning, Tan-RSS-18}. 
%
In particular, policies modeled by deep networks are subject to spectral biases \cite{ronen2019convergence,battaglia2018relational, bietti2019inductive} that make them more likely to learn and generalize the low-frequency patterns that exist in the control trajectories.
Hence, choosing the right action space in which to define these trajectories remains a widely recognized problem in both reinforcement learning \cite{peng2017learning} as well as imitation learning, where demonstrations can be provided in a wide range of formats -- e.g., continuous teleoperation in Cartesian or joint space, kinesthetic teaching, etc. -- each changing the underlying characteristics of the training data.

\begin{figure}[t!]
\centering
  \vspace{0.5em}
  \noindent\includegraphics[width=1.0\linewidth]{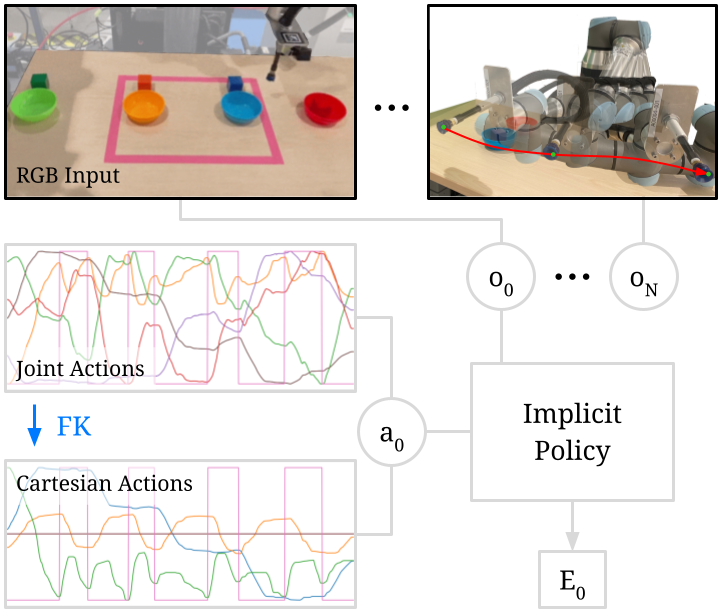} 
  \caption{Enabled by an implicit policy formulation, Implicit Kinematic Policies (IKP) provide both Joint and Cartesian action representations (linked via forward kinematics) to a model that can pick up on inductive patterns in both spaces. Subsequently, this allows our end-to-end policies to extract the best action space for a wide variety of manipulation tasks ranging from tabletop block sorting to whole-arm sweeping.}
  \label{fig:teaser}
  \vspace{-1.5em}
\end{figure}

Although considerable research has been devoted to finding the right action space for a given application \cite{peng2017learning}, much less attention has been paid to figuring out how our models could instead discover for themselves which optimal combination of action spaces to use. 
This goal remains particularly unclear for most \textit{explicit} policies $f_\theta(\textbf{o})=\hat{\textbf{a}}$ with feed-forward models that take as input observations $\textbf{o}$ and output actions $\textbf{a}\in\mathcal{A}$. Explicit formulations must either specify a single action space to output from the model and potentially convert predictions into another desired space (e.g. with inverse kinematics), or use multiple action spaces with multiple outputs and losses which can be subject to conflicting gradients \cite{yu2020gradient} from inconsistent predictions \cite{martin2019iros}.


In this work, we demonstrate that we can train deep policies to learn which action space to use for imitation, made possible by implicit behavior cloning (BC) \cite{florence2021implicit}. In contrast to its explicit counterpart, \textit{implicit} BC formulates imitation as the minimization of an energy-based model (EBM) $\hat{\textbf{a}}=\arg\min_{\textbf{a} \in \mathcal{A}} E_{\theta}(\textbf{o},\textbf{a})$ \cite{lecun2006tutorial}, which regresses the optimal action via sampling or gradient descent \cite{welling2011bayesian, du2019implicit}. Since actions are now inputs to the model, we propose presenting the model with the same action represented in multiple spaces while remaining consistent between each other, allowing the model to pick up on inductive patterns \cite{battaglia2018relational, bietti2019inductive} from all action representations.

We study the benefits of this multi-action-space formulation with Cartesian task and joint action spaces in the context of learning manipulation skills. Since both spaces are linked by the kinematic chain, we can integrate a differentiable kinematic module within the deep network -- a formulation which we refer to as Implicit Kinematic Policies (IKP). IKP can be weaved into an implicit autoregressive control strategy introduced in \cite{florence2021implicit}, where each action dimension is successively and uniformly sampled at a time and passed as input to the model. This exposes the model to not only the joint configuration and end-effector pose, but also the Cartesian action representations of {\em{every rigid body link}} of the arm as input. This can provide downstream benefits when learning whole-body manipulation tasks that may have emergent patterns in the trajectory of any given link of the robot. Furthermore, a key aspect of Implicit Kinematic Policies is that in addition to exposing the implicit policy to both Cartesian and joint action spaces, it also enables incorporating learnable layers throughout the kinematic chain, which can be used to optimize for residuals in either space.

Our main contribution is Implicit Kinematic Policies, a new formulation that integrates forward kinematics as a differentiable module within the deep network of an autoregressive implicit policy, exposing both joint and Cartesian action spaces in a kinematically consistent manner. Behavior cloning experiments across several complex vision-based manipulation tasks suggest that IKP, without any modifications, is broadly capable of efficiently learning both prehensile and non-prehensile manipulation tasks, even in the presence of miscalibrated joint encoders -- results that may pave the way for more data efficient learning on low-cost or cable-driven robots, where low-level joint encoder errors (due to drift, miscalibration, or gear backlash) can propagate to large non-linear artifacts in the Cartesian end effector trajectories. IKP achieves 85.9\%, 97.5\% and 92.4\% on a sweeping, non-prehensile box lifting and precise insertion task respectively -- performance that is on par with or exceeds that of standard implicit or explicit baselines (using the best empirically chosen action space per individual task). We also provide qualitative experiments on a real UR5e robot suggesting IKP's ability to learn from noisy human demonstrations in a data-efficient manner.
Our experiments provide an initial case study towards more general action-space-agnostic architectures for end-to-end robot learning.

\section{Related Work}
\subsection{Learning Task-Specific Action Representations}
While much of the early work in end-to-end robot learning has focused on learning explicit policies that map pixels to low-level joint torques \cite{ddpg2016, levine_finn_2016}, more recent work has found that the choice of action space can have a significant impact on downstream performance for a wide range of robotic tasks ranging from manipulation to locomotion \cite{martin2019variable, duan2021learning, peng2017learning, Viereck_2018, varin2019comparison}. In the context of deep reinforcement learning for locomotion, \cite{peng2017learning} find that using PD joint controllers achieves higher sample efficiency and reward compared to torque controllers on bipedal and quadrupedal walking tasks in simulation, whereas \cite{Viereck_2018} interestingly find that torque controllers outperform PD joint controllers for a real robot hopping task. In contrast, \cite{duan2021learning} shows that using task space actions are more effective for learning higher level goals of legged locomotion compared to joint space controllers and demonstrate results on a real bipedal robot.

Careful design of the action space is equally important when learning manipulation skills with high DoF robots \cite{martin2019variable, varin2019comparison}. Prior approaches successfully use direct torque control or PD control with joint targets to learn a variety of tasks such as pick-and-place, hammering, door opening \cite{rajeswaran2018learning} and stacking \cite{popov2017dataefficient}, but these methods often require lots of data and interaction with the environment to generalize well and are also sensitive to feedback gains in the case of PD control.

Other works default to low-level residual cartesian PD controllers in order to learn prehensile tasks such as block stacking \cite{johannink2018residual} and insertion \cite{silver2019residual}, but \cite{martin2019variable, varin2019comparison} instead vouch for a task-space impedance controller to support more compliant robot control. However, this approach is agnostic to the full joint configuration of the robot and focuses on tabletop manipulation which is generally better defined in task space. Fundamentally, these works share the common theme of choosing the single best action space for a particular task. In this work, we are more concerned with the problem of extracting the best combination of action spaces.



\subsection{Difficulties in Learning High-Frequency Functions}
The proposed benefits of implicit kinematic policies rely on the hypothesis that simpler patterns are easier for deep networks to learn. Although theoretical analyses of two-layer networks and empirical studies of random training labels have shown that deep networks are prone to memorization \cite{Zhang2017UnderstandingDL}, further studies on network convergence behavior and realistic dataset noise have found that this result does not necessarily explain performance in practice. Instead, deep networks first learn patterns across samples before memorizing data points \cite{Arpit2017ACL}. \cite{Rahaman2019Spectral} formalize this finding and show that deep networks have a learning bias towards low-frequency functions by using tools from Fourier analysis and \cite{Basri2019TheCR} extend this result by showing that networks fit increasingly higher frequency functions over the course of training. Within learning-based robotics, multiple works have found the related result that higher-level action spaces lead to improved learnability. Specifically, the frequency of the action space can have an outsized effect on the robustness and quality of learned policies \cite{peng2017learning}. This could explain why learned, latent actions improve performance and generalizability for policies trained either with reinforcement learning \cite{Haarnoja2018LatentSP,Hausman2018LearningAE,Nair2020HierarchicalFS,Pflueger2020PlanSpaceSE} or imitation learning \cite{fox2017multi}.



\begin{figure}[t]
\centering
  \noindent\includegraphics[width=\linewidth]{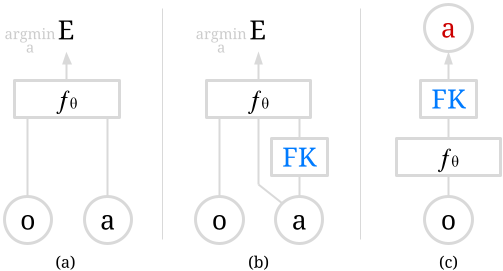} 
  \caption{Integrating forward kinematics (FK), in particular, synergizes with implicit policies (a, b), where actions $\textbf{a}$ are sampled in joint space, and both joints and corresponding Cartesian actions along with observations $\textbf{o}$ can be fed as inputs to the EBM $f_\theta$. For explicit policies (c), it becomes less clear how to expose both action spaces in a kinematically consistent way to the deep network, except via predicting joints from observations, then using forward kinematics to backpropagate gradients from losses imposed on the final output Cartesian actions (in red). In this work, we investigate both formulations and find that implicit (b) yields better performance.
  }
  \label{fig:key-idea}
  \vspace{-2em}
\end{figure}

\section{Background}
\subsection{Problem Formulation}
We consider the imitation learning setting with access to a fixed dataset of observation-action pairs. We frame this problem as supervised learning of a policy $\pi: \mathcal{O} \rightarrow \mathcal{A}$ where $\mathbf{o} \in \mathcal{O} = \mathbb{R}^m$ represents an observational input to the policy and $\mathbf{a} \in \mathcal{A} = \mathbb{R}^d$ represents the action output. The policy $\pi$, with parameters $\theta$, can either be formulated as an {\em{explicit}} function $f_{\theta}(\mathbf{o}) \rightarrow \hat{\mathbf{a}}$ or as an {\em{implicit}} function $\arg\min_{\mathbf{a}\in\mathcal{A}} E(\mathbf{o}, \mathbf{a}) \rightarrow \hat{\mathbf{a}}$. In this work, we adopt the implicit formulation and build upon recent work which we describe in more detail in Section \ref{implicitbcsec}. We operate in the continuous control setting, where our policy infers a desired target configuration at 10 Hz, and a low-level controller asynchronously reaches desired configurations with a joint-level PD controller. Our $\mathcal{O}$ contains both image observations and proprioceptive robot state (i.e. from joint encoders).

\subsection{Implicit BC}
\label{implicitbcsec}
We use an energy-based, contrastive loss to train our implicit policy, modeled with the same late-fusion deep network architecture as in \cite{florence2021implicit} with 26 convolutional ResNet layers \cite{resnets_2016} for the image encoder and 20 dense ResNet layers for the EBM. Specifically, to optimize $E_{\theta}(\cdot)$, we train an InfoNCE style loss \cite{Oord2018RepresentationLW} on observation-action samples \cite{florence2021implicit}. Our dataset consists of $\{\mathbf{o}_i, \mathbf{a}^*_i\}$ for $\mathbf{o}_i \in \mathbb{R}^m$ and $\mathbf{a}_i \in \mathbb{R}^d$ and from regression bounds $\mathbf{a}_{\min} , \mathbf{a}_{\max} \in \mathbb{R}^d$ we generate negative samples $\{\tilde{\mathbf{a}}_i^j\}_{j=1}^{N_{\text{neg.}}}$. The loss equates to the negative log likelihood of $p(\mathbf{a}|\mathbf{o})$ where we use negative samples to approximate the normalizing constant:
\begin{align*}
    \mathcal{L}_{\text{InfoNCE}} &= \sum_{i=1}^N -\log(\tilde{p}_\theta(\mathbf{a}^*_i|\mathbf{o}_i,\{\tilde{\mathbf{a}}_i^j\}_{j=1}^{N_{\text{neg.}}})) \\
    \tilde{p}_\theta(\mathbf{a}^*_i|\mathbf{o}_i,\{\tilde{\mathbf{a}}_i^j\}_{j=1}^{N_{\text{neg.}}}) &= \frac{e^{-E_\theta(\mathbf{o}_i, \mathbf{a}^*_i)}}{e^{-E_\theta(\mathbf{o}_i, \mathbf{a}^*_i)} + \sum_{j=1}^{N_{\text{neg.}}}e^{-E_\theta(\mathbf{o}_i, \tilde{\mathbf{a}}_i^j)}}
\end{align*}
To perform inference, given some observation $\mathbf{o}$, we minimize our learned energy function, $E_{\theta}(\mathbf{o}, \mathbf{a})$,  over actions in $\mathcal{A}$ using a sampling-based optimization procedure.


\label{implicitbc}
\label{background}

\section{Method}

Here we present the details of our proposed Implicit Kinematic Policies (IKP) method.  First, we will present our extension to the implicit policy formulation which provides multiple action space representations as input. (Sec.~\ref{subsec:multi-action-space}). Sec.~\ref{subsec:multi-action-space-explicit} also compares the implicit multi-action-space formulation to its explicit counterpart and discusses the relevant tradeoffs. We then describe how to perform autoregressive training and inference with implicit policies in a way that exposes the action trajectories of every joint and link in the robot to the model (Sec.~\ref{subsec:autoregressive-formulation}). Finally, we discuss a motivating application of controlling miscalibrated robots for precise tasks, and how IKP is uniquely suited to automatically compensate for this noise through the use of strategically placed residual blocks in the model (Sec.~\ref{subsec:residual-formulation}).


\subsection{Multi-Action-Spaces With Implicit Policies}\label{subsec:multi-action-space}

Our formulation enables the implicit policy $\hat{\bfa} = \arg\min_{\bfa} E_{\theta}(\bfo, \bfa)$ to have access to multiple action spaces, which are constrained to be consistent. Specifically, our implicit multi-action-space formulation is of the form:
\vspace{-.85em}

\begin{equation}
\begin{aligned}
\underset{\bfa, \bfa', ...}{\arg\min} \quad & E_{\theta}(\bfo, \bfa, \bfa', ...)\\
\textrm{s.t.} \quad & \bfa = \mathcal{T}(\bfa'), ...
\end{aligned}
\label{eq:general-multi-action-space}
\end{equation}
where $\bfa \in \mathcal{A}$ is one parameterization of the action space, $\bfa' \in \mathcal{A}'$ is a different parameterization of the action space, and  $\mathcal{T}(\cdot): \mathcal{A}' \rightarrow \mathcal{A}$ is a transformation between the two action spaces. Of course, $N$ different consistent parameterizations of the action space could be represented, which is depicted in Eq.~\ref{eq:general-multi-action-space} by the ellipses (...).

In particular for robots, we are interested in representing both {\em{joint-space}} and {\em{cartesian-space}} actions, which is a case of Eq.~\ref{eq:general-multi-action-space} in which the transformation between the two action spaces is forward kinematics (FK):
\vspace{-0.75em}
\begin{equation}
\begin{aligned}
\underset{\bfa_{\text{joints}}, \ \bfa_{\text{cartesian}}}{\arg\min} \quad & E_{\theta}(\bfo, \bfa_{\text{joints}}, \bfa_{\text{cartesian}})\\
\textrm{s.t.} \quad & \bfa_{\text{cartesian}} = FK(\bfa_{\text{joints}})
\end{aligned}
\label{eq:joints-cartesian-implicit}
\end{equation}
Specifically, we accomplish this by first sampling input actions in joint space $\bfa_{\text{joints}} \in \mathcal{A}_{\text{joints}}$, then computing the corresponding Cartesian (task) space actions via forward kinematics ($FK$), $\bfa_{\text{cartesian}} = FK(\bfa_{\text{joints}})$, and finally concatenating and passing the combined representation into the model $E_{\theta}(\bfo, \bfa_{\text{joints}}, \bfa_{\text{cartesian}})$. A visualization of this model is shown in the middle portion of Fig.~\ref{fig:key-idea}. Our hypothesis is that the model can use this redundant action representation to exploit patterns in both spaces, akin to {\em{automatically}} discovering the best combination of action spaces -- this hypothesis will be tested in our Experiments section.
As we will show in Sec.~\ref{subsec:autoregressive-formulation}, we can perform training and inference for $E_{\theta}$ through using autoregressive derivative-free optimization.

\begin{figure*}[t!]
\label{fkdiagram}
\centering
  \noindent\includegraphics[width=\textwidth]{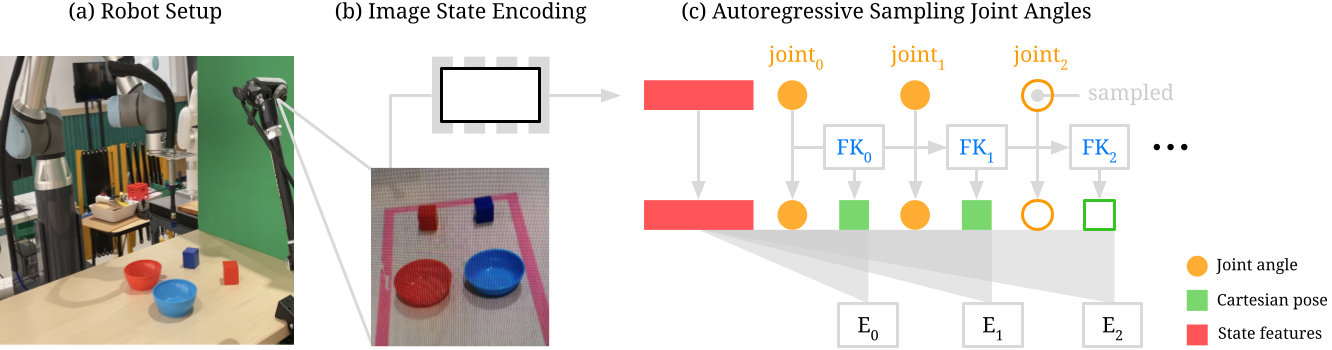} 
  \caption{ \textbf{Method overview.} Given an image captured from an RGB camera overlooking the robot workspace (a), we feed it as input to a deep convolutional network (b) to get a latent state representation. We then predict desired robot joint actions by leveraging implicit autoregression \cite{florence2021implicit, nash2019autoregressive} (c) with state-conditioned EBMs to progressively sample each action dimension (joint angle) at a time: \ie we uniform sample $\text{j}_n$, feed it to $\text{FK}_n$ (forward kinematics to link $n$, prepended with deep layers) to get the Cartesian pose $\text{C}_{n}$, which is then concatenated with the latent state and all previously sampled argmin joint dimensions $[\text{j}_n, \text{j}_{n-1}, ..., \text{j}_0]$ and Cartesian representations $[\text{C}_{n-1}, ..., \text{C}_0]$ and fed to an 8-layer EBM $\text{E}_{n}$ to compute the argmin over $\text{j}_n$.
  }
  \label{fig:method-overview}
  \vspace{-1.5em}
\end{figure*}

\subsubsection{Multi-Action-Spaces With Explicit Policies}
\label{subsec:multi-action-space-explicit}
Consider the above formulation in contrast to an explicit policy, $\hat{\bfa}=f_{\theta}(\bfo)$, where the mapping $f_{\theta}(\cdot)$ maps to one specific action space.  It is possible to provide different losses on different transformations of the explicit policy's action space, for example $\mathcal{L}_{\text{joints}}(\hat{\bfa}, \bfa^*)$ and $\mathcal{L}_{\text{cartesian}}(FK(\hat{\bfa}), FK(\bfa^*))$, where $FK$ represents differentiable forward kinematics. However, this can lead to conflicting gradients and requires choosing relative weightings, $\lambda$, between these losses: $\mathcal{L}_{\text{total}} = \mathcal{L}_{\text{joints}} + \lambda \mathcal{L}_{\text{cartesian}}$. This issue of loss balancing is exacerbated if the explicit policy were to regress the Cartesian representation of every link in the robot in addition to the joint configuration and end-effector pose. Additionally, the explicit policy predicts a single joint configuration even though there may be multiple correct configurations when using kinematically redundant robots. It is possible to first regress Cartesian actions and subsequently recover the corresponding joint configuration via inverse kinematics, however, IK solvers are only approximately differentiable which poses challenges for end-to-end training. A visualization of this explicit policy is shown in the rightmost column of Fig.~\ref{fig:key-idea}.

\subsection{Autoregressive Implicit Training and Inference}
\label{subsec:autoregressive-formulation}
As in \cite{florence2021implicit}, we employ an autoregressive derivative-free optimization method for both training and inference.  Sampling-based, derivative-free optimization synergizes with the multi-action-space constrained energy model (Eq.~\ref{eq:general-multi-action-space}), since gradient-based optimization (i.e. through Langevin dynamics \cite{florence2021implicit}), while possible, would require a solution to synchronize between the different action spaces.\footnote{Specifically, when using the implicit model gradients, with learning rate $\lambda$, the actions are updated as $\bfa_{\text{joints}}^{k+1} = \bfa_{\text{joints}}^{k} - \lambda \nabla_{\bfa_{\text{joints}}} E_{\theta}(\cdot)$ and $\bfa_{\text{cartesian}}^{k+1} = \bfa_{\text{cartesian}}^{k} - \lambda \nabla_{\bfa_{\text{cartesian}}} E_{\theta}(\cdot)$.  The updated actions 
actions may not be consistent with each other -- even for an analytical model, there will be differences in the first-order derivatives of the different spaces.}

The autoregressive procedure uses $m$ models, one model $E_{\theta}^{j}(\mathbf{o}, \bfa^{0:j})$ for each dimension $j = 1, 2, ...,m$. In contrast with prior work \cite{florence2021implicit}, instead of $\bfa$ only representing a single action space, our proposed method represents, for each $j$th model, both the 6DoF Cartesian pose of the $j$th link {\em{and}} the joint of that link into $\bfa$. Similarly in $\mathbf{o}$ we also represent the dual Cartesian-and-joint state of each link.  Since our method strictly samples actions in joint space and the state is represented by the joint configuration, the Cartesian representations for both the state $\bfo$ and actions $\bfa$ are acquired through $FK(\bfo)$ and $FK(\bfa)$ for every dimension $j = 1, 2, ...,m$. We visualize this procedure in Fig. \ref{fig:method-overview}.

\subsection{Residual Forward Kinematics}
\label{subsec:residual-formulation}
In robotics, we often assume access to accurate kinematic descriptions and joint encoders, but this assumption can be a significant source of error in the context of low-cost or cable-driven robots. In these settings, the errors often take the form of fixed, unknown linear offsets in each joint motor encoder or link representation. If calibration isn't performed, both cases can cause heavy non-linear offsets in end-effector space which can drastically affect the performance on high-precision tasks. By adding dense residual blocks both before and after joint actions are sampled in the autoregressive procedure, we can encourage our model to learn these linear offsets at each joint and link rather than solving the more difficult problem of learning a highly non-linear offset in task space. This extension is only possible due to the full differentiability of the forward kinematics layer, which allows gradients to flow through the residual blocks. Concretely, we redefine our multi-action-space formulation with residuals as follows:
\vspace{-1.5em}

\begin{equation}
\begin{aligned}
\underset{\bfa_{\text{joints}}, \ \bfa_{\text{cartesian}}}{\arg\min} \quad & E_{\theta}(\bfo, \bfa_{\text{joints}}, \bfa_{\text{cartesian}})\\
\textrm{s.t.} \quad & \bfa_{\text{cartesian}} = FK\big(\bfa_{\text{joints}} + \Delta_{\theta}(\bfa_{\text{joints}})\big)
\end{aligned}
\label{eq:joints-cartesian-residuals-implicit}
\end{equation}
where $\Delta_{\theta}(\bfa_{\text{joints}}) \in \mathcal{A}_{\text{joints}}$ is a new learnable module in the implicit model which gives the model the inductive bias that there may be imperfect calibration of the joint measurements (i.e., biased joint encoder measurements).  The module $\Delta_{\theta}(\cdot)$ can be learned during training and we hypothesize that the EBM may be able to automatically learn the $\Delta_{\theta}(\cdot)$ which provides the lowest-energy fit of the data. This can be interpreted as automatically calibrating the biased joint encoders. Experiments testing this hypothesis are in Section~\ref{subsec:residual-experiments}.

\begin{figure}[t]
\centering
    \noindent\includegraphics[width=\linewidth]{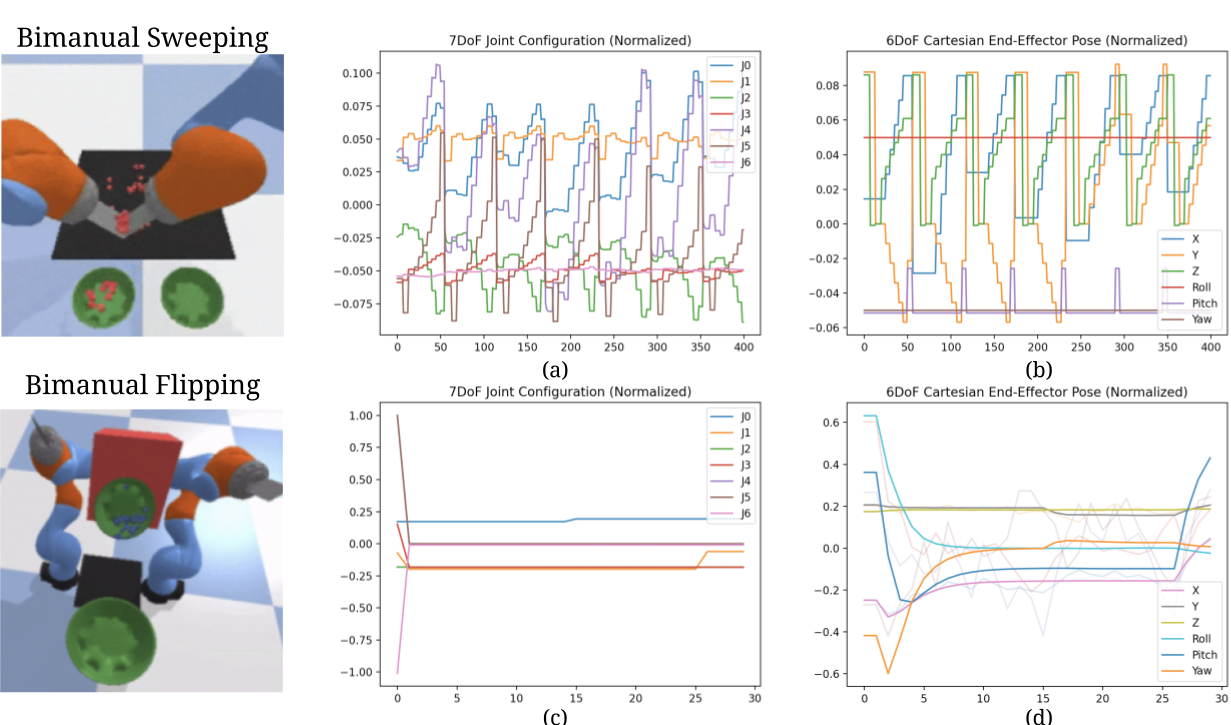}
  \caption{Expert trajectories for bimanual sweeping are characterized by distinct patterns in Cartesian space (b), \eg y-values experience a mode-change in the latter half of the episode when the policy switches bowls in which to drop particles. Such patterns are less salient in joint space (a). The opposite is true for bimanual flipping, where linear patterns emerge in joint space (c), whereas they are less salient in Cartesian space (d) especially with randomized end effector poses (semi-transparent plots).}
  \label{fig:tasks}
  \vspace{-1.5em}
\end{figure}

\section{Experiments}
We evaluate IKP across several vision-based continuous control manipulation tasks with quantitative experiments in simulation, as well as qualitative results on a real robot. All tasks require generalization to unseen object configurations at test time. The goals of our experiments are two-fold: (i) across tasks where one action space substantially outperforms the other, we investigate whether IKP can achieve the best of both and perform consistently well across all tasks, and (ii) given a miscalibrated robot with unknown offsets in the joint encoders (representing miscalibration or low-cost encoders), we study whether IKP can autonomously learn to compensate for these offsets while still succeeding at the task. Across all experiments, a dataset of expert demonstrations is provided by a scripted oracle policy (in simulation), or by human teleoperation (in real).


\subsection{\textbf{Simulated Bimanual Sweeping and Flipping}}
In a simulated environment, we evaluate on two bimanual tasks -- sweeping and flipping -- both of which involve two 7DoF KUKA IIWA robot arms equipped with spatula-like end-effectors positioned over a $0.4m^2$ workspace. The setup for sweeping is identical to that presented in Florence et al. \cite{florence2021implicit}, where given a pile of 50-60 particles on the workspace, the task is to scoop and evenly distribute all particles into two bowls located next to the workspace. For flipping, given a bowl of 20-30 particles attached to the top of a $0.2m^3$ box, the task is to flip the box to pour the particles into a larger bowl positioned near the base of the arms. Both tasks are designed to involve significant coordination between the two arms. During sweeping, for example, scooping up particles and transporting them to the bowls requires carefully maintaining alignment between the tips of the spatulas to avoid dropping particles. Flipping, on the other hand, requires aligning the elbow surfaces against the sides of the box, and using friction to carry the box as it pours the particles into the bowl -- any subtle misalignment of the elbow against the sides can lead to the box slipping away or tipping over. For both tasks, we generate fixed datasets of 1,000 demonstration episodes using a scripted oracle with access to privileged state information, including object poses and contact points, which are not accessible to policies.

We train two Implicit BC \cite{florence2021implicit} baselines \textit{without} Residual Kinematics: one using a 12DoF Cartesian action space (6DoF end-effector pose for each arm), and another using a 14DoF joint action space (7DoF for each arm). From the quantitative results presented in Tab. \ref{table:bimanual-experiments}, we observe that the performance of Implicit BC on bimanual sweeping degrades substantially when using a joint action space. This is likely that learning the task (\i.e. distribution of oracle demonstrations) involves recognizing a number of Cartesian space constraints that are less salient in joint space: \eg keeping the Cartesian poses of the spatulas aligned with each other while scooping and transporting particles, and maintaining z-height with the tabletop.
For bimanual flipping, on the other hand, we observe the opposite: where the performance of Implicit BC performs well with joint actions, but poorly with Cartesian actions. We conjecture that the task involves more whole-body manipulation, where fitting policies in joint space are more likely to result in motion trajectories that accurately conform to the desired elbow contact with the box. 
\vspace{-.7em}
\begin{table}[h]
  \centering
  \caption{\scriptsize Performance Measured in Task Success (Avg $\pm$ Std \% Over 3 Seeds)}
  \begin{tabular}[b]{@{}lcccccccc@{}}
  \toprule
  Task & Sweeping & Flipping \\
  {\em{Oracle's action space}} & \em{cartesian} & \em{joints} \\
  Policy Action Space &  &  \\
  \midrule
  Cartesian & 79.4 $\pm$ 2.1  & 38.6 $\pm$ 5.2\\
  Joints & 44.3 $\pm$ 3.2 & \textbf{98.4 $\pm$ 1.4} \\
  Joints + Cartesian (Ours)    & \textbf{85.9 $\pm$ 1.5} &  \textbf{97.5 $\pm$ 1.2} \\
  \bottomrule
  \end{tabular}
  \vspace{-1.0em}
  \label{table:bimanual-experiments}
 
\end{table}

Our proposed method, Implicit Kinematic Policies (labeled as Joints + Cartesian in Tab. \ref{table:bimanual-experiments}) leverages both action spaces. Results suggest its performance is not only on par with the best bespoke action space for the task, but also surprisingly exceeds the performance of Cartesian actions for bimanual sweeping. Upon further inspection of the action trajectories for bimanual sweeping, it is evident that the Cartesian space action trajectories (Fig. \ref{fig:tasks}b) are lower frequency and contain more piecewise linear structure in comparison to the joint space action trajectories (Fig.~\ref{fig:tasks}a). Implicit policies have shown to thrive in conditions where this structure is present \cite{florence2021implicit}. In contrast, for bimanual flipping, the joint space action trajectories (Fig.~\ref{fig:tasks}c) appear much more linear and low frequency than the corresponding arc-like Cartesian trajectories (Fig.~\ref{fig:tasks}d). We also add small perturbations to the end-effector pose highlighted by the semi-transparent lines in plot (d) of Fig.~\ref{fig:tasks}. Note that the Implicit (Cartesian) policy uses a generic IK solver which can return multiple distinct joint target commands, causing less stable trajectories overall.

\begin{figure}[t]
\centering
  \noindent\includegraphics[scale=0.2]{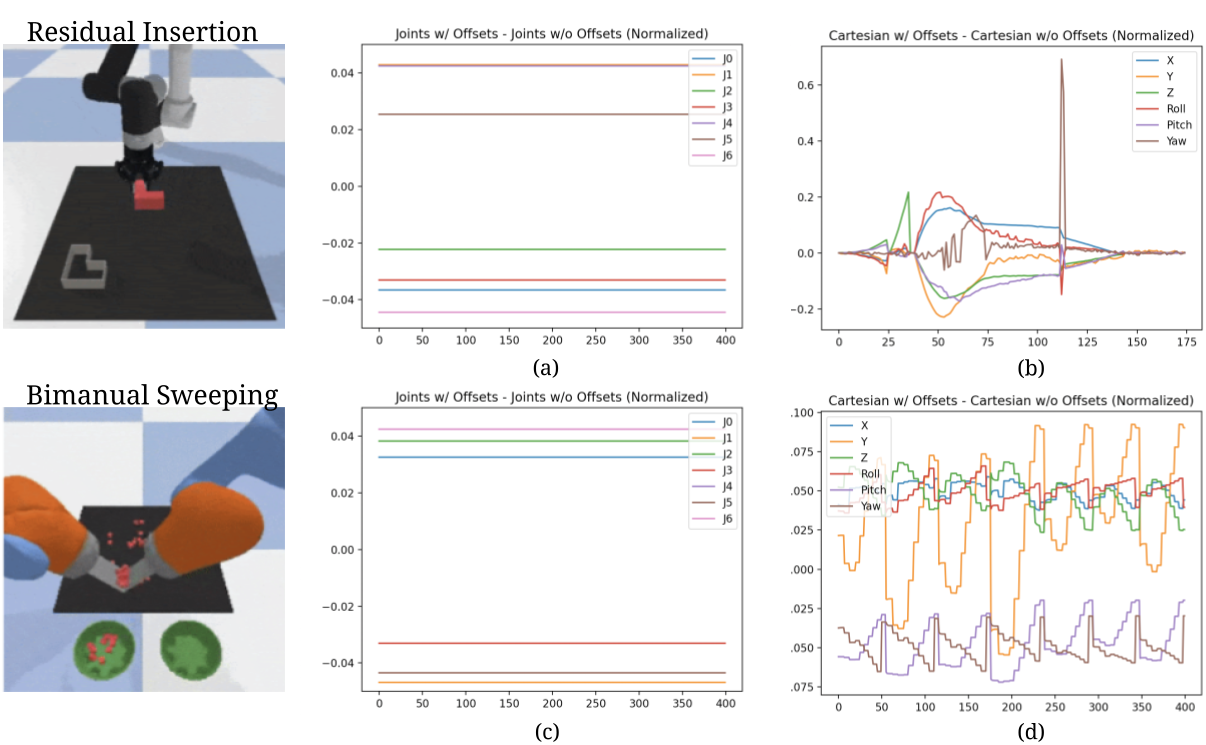} 
  \vspace{-0.5em}
  \caption{Simple constant joint encoder errors (which may appear from drift in low-cost or cable-driven robots) can propagate to non-linear offsets in Cartesian space. By learning joint space residuals to compensate for these offsets, IKP is better equipped to generalize over such errors than standard end-to-end policies trained in Cartesian space for tabletop manipulation.}
  \label{fig:residual-deltas}
  \vspace{-1.5em}
\end{figure}

\subsection{\textbf{Simulated Miscalibrated Sweeping and Insertion}}
\label{subsec:residual-experiments}

We design two additional tasks to evaluate IKP's ability to learn high precision tasks in the presence of inaccurate joint encoders, described in Sec.~\ref{subsec:residual-formulation}. To replicate such a scenario, we simulate these errors by adding constant $<$ 2 degree offsets to each of the six revolute joints on a UR5e robot. These offsets are visualized and described in more detail in Fig.~\ref{fig:residual-deltas}. We test the effect of these offsets on the bimanual sweeping task and a new insertion task where the goal is to insert an L-shaped block into a tight fixture. Demonstrations for both tasks are provided in the form of a Cartesian scripted oracle with privileged access to the underlying joint offsets and can compensate for them. This is akin to a human using visual servoing to compensate for inaccurate encoders when teleoperating a real robot to collect expert demonstrations. Although both tasks are generally well-defined in Cartesian space as a series of linear step functions, the joint offsets induce high-frequency non-linear artifacts in the end-effector trajectory (Fig.~\ref{fig:residual-deltas}b and \ref{fig:residual-deltas}d) causing poor performance with the Implicit (Cartesian) policy (results in Tab.~\ref{table:noisy-experiments}). Poor performance persists for miscalibrated insertion despite using 10x the data. Alternatively, even though the encoder offsets only cause linear shifts in the joint trajectories (Fig.~\ref{fig:residual-deltas}a and \ref{fig:residual-deltas}c), both tasks are less structured in joint space and as a result Implicit (Joints) performs significantly worse on miscalibrated bimanual sweeping and comparatively worse on miscalibrated insertion. Implicit (Joints) performs relatively well on the miscalibrated insertion task due to the simplicity of the pick and place motion, but struggles to generalize when the block nears the edge of workspace as the joint trajectories become increasingly non-linear in those regions. The performance only slightly improves when using 1000 demonstrations. IKP provides a best-of-all-worlds solution by learning the linear offsets in joint space through the residual blocks while simultaneously exploiting the unperturbed Cartesian trajectories through forward kinematics on the shifted joint actions to generalize. IKP achieves the highest performance on miscalibrated insertion with both 100 and 1000 demonstrations, and significantly higher performance on miscalibrated bimanual sweeping. Interestingly, not only do Explicit and ExplicitFK perform significantly worse than both Implicit (Joints) and IKP for both tasks, but ExplicitFK also provides little to no additional benefit over Explicit, even in the presence of more data.


\begin{table}[h]
  \centering
  \caption{\scriptsize Miscalibrated Joint Encoder Experiments, Performance Measured in Task Success (Avg $\pm$ Std. \% Over 3 Seeds). (``J+C'') is short for Joints+Cartesian.}
  \begin{tabular}[b]{@{}lcccccccc@{}}
  \toprule
  Task & Sweeping & \multicolumn{2}{c}{Block Insertion}\\
  {\em{Oracle's action space}} & \em{cartesian} & \multicolumn{2}{c}{\em{cartesian}}\\
  Method (Action Space) & 1000 & 100 & 1000\\
  \midrule
  Explicit (Joints)               & 38.2 $\pm$ 3.4 & 73.8 $\pm$ 4.3 & 74.2 $\pm$ 2.5\\
  ExplicitFK (J + C) (Ours)               & 40.3 $\pm$ 4.6 & 72.4 $\pm$ 3.8 & 75.4 $\pm$ 2.8\\
  Implicit (Cartesian)               & 3.1 $\pm$ 2.2 & 0.0 $\pm$ 0.0 & 0.0 $\pm$ 0.0\\
  Implicit (Joints)                    & 46.3 $\pm$ 1.8 & 82.3 $\pm$ 2.9 & 85.2 $\pm$ 3.1 \\
  IKP (J + C) (Ours)  & \textbf{84.5 $\pm$ 1.2} & \textbf{88.8 $\pm$ 3.4} & \textbf{92.4 $\pm$ 2.6}\\
  \bottomrule
  \end{tabular}
  \vspace{-1.0em}
  \label{table:noisy-experiments}

\end{table}

\subsection{\textbf{Real Robot Sorting, Sweeping, and Alignment}}
\begin{figure}[t]
\centering
  \noindent\includegraphics[scale=0.23]{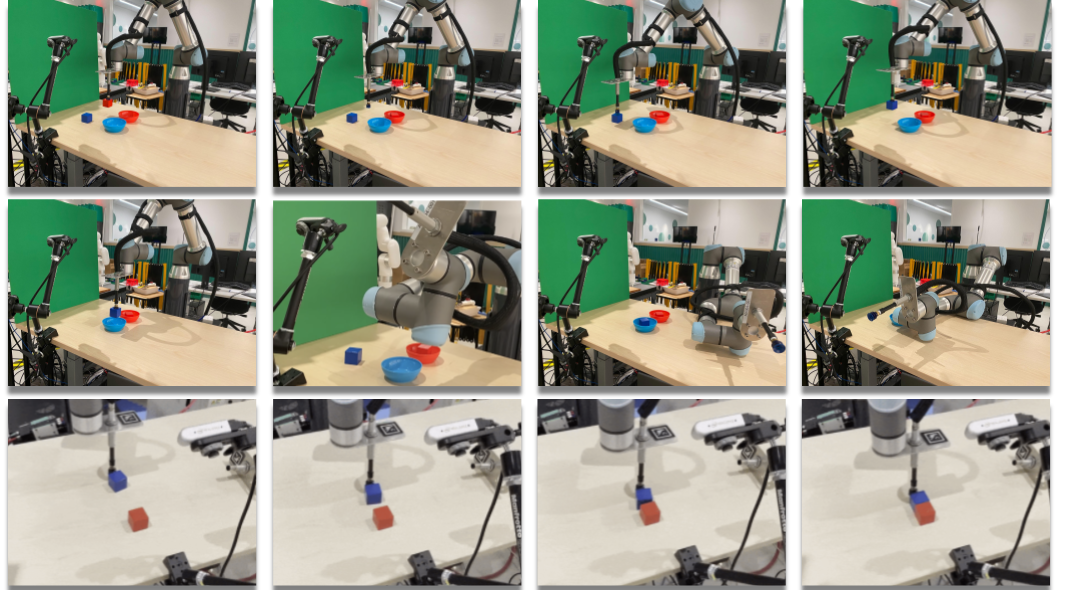} 
  \caption{Real robot IKP rollouts on a sweeping and sorting task (top two rows) and a block alignment task (bottom row).}
  \label{fig:real-tasks}
  \vspace{-2em}
\end{figure}

We conduct qualitative experiments with a real UR5e robot on two tasks: 1) sorting two blocks into bowls and subsequently sweeping the bowls, and 2) aligning a red block with a blue block, where both tasks have initial locations randomized). Our goal with these experiments is two-fold: i) to demonstrate that IKP can run on real robots with noisy human continuous teleop demonstrations with as few as 100 demonstrations, and ii) to show that Implicit Kinematic Policies can perform both whole-body and prehensile tasks. Demonstrations for the block sorting portion of the first task are provided using continuous Cartesian position control based teleoperation at 100 Hz with a 3D mouse. Once the blocks are sorted, the teleoperator switches to a joint space PD controller with the same 3D mouse to guide the arm into an extended pose in order to sweep the bowls. For the alignment task, demonstrations are also provided using x, y and z Cartesian PD control at 100 Hz. Both tasks take 480${\times}$640 RGB images (downsampled to 96${\times}$96) from an Intel RealSense D435 camera at a semi-overhead view as input to the policy, and no extrinsic camera calibration is used. We show successful rollouts for both tasks in Fig.~\ref{fig:real-tasks}. 

\section{Discussion}
In future work, we will investigate using the multi-action-space formulation to extend beyond joint and cartesian PD control by incorporating the forward and/or inverse robot dynamics into the network which will allow us to expose joint torques and velocities to our model in a fully differentiable way. We hypothesize that this additional information will be particularly helpful in dynamic manipulation tasks and visual locomotion where torque and velocity action trajectories may appear more structured. We would also like to further test our residual framework within IKP on robots that have non-linear drift in joint space which may represent a larger set of low-cost robots. Finally, we would like to utilize our fully differentiable residual forward kinematics module to learn the link parameters themselves, which can be a promising direction for controlling soft and/or continuum robots.

\section*{Acknowledgments}

\small The authors would like to thank Vincent Vanhoucke, Vikas Sindhwani, Johnny Lee, Adrian Wong, Daniel Seita and Ryan Hoque for helpful discussions and valuable feedback on the manuscript.



\bibliographystyle{IEEEtranS}
\bibliography{main}

\end{document}